# Vision-based Pose Estimation for Augmented Reality : A Comparison Study

Hayet Belghit[1], Abdelkader Bellarbi[2], Nadia Zenati[3], Samir Otmane[4]

[1,2,3] CDTA, Baba Hassen, Algiers, Algeria.

[4] IBISC, Univ Evry, Université Paris-Saclay, 91025, Evry, France.

[1] hbelghit@cdta.dz, [2] abellarbi@cdta.dz, [3] nzenati@cdta.dz, [4] samir.otmane@ibisc.univ-evry.fr

*Abstract*—**Augmented reality aims to enrich our real world by inserting 3D virtual objects. In order to accomplish this goal, it is important that virtual elements are rendered and aligned in the real scene in an accurate and visually acceptable way. The solution of this problem can be related to a pose estimation and 3D camera localization. This paper presents a survey on different approaches of 3D pose estimation in augmented reality and gives classification of key-points-based techniques. The study given in this paper may help both developers and researchers in the field of augmented reality.**

*Keywords—Augmented Reality, Computer Vision, Pose Estimation, Descriptors.*

## I. INTRODUCTION

Augmented reality (AR) is the technology that enhance human's real-world perception with computer generated elements by superimposing the virtual world on the real world.

The first AR interface was developed by Sutherland in the1960's [1], but it was 30 years later that the term augmented reality was first used in 1992 by Thomas Caudell and David Mizell [2] to describe a semi-transparent helmet, used by aeronautical electricians and visualizing virtual information on real images. Nowadays, AR is becoming popular, and it is used in many applications [3][4][5][6].

A lot of definitions have been then given to Augmented Reality. Each one defined it according to a specific aspect [7][8][9][10]. However, most of these definitions mentioned that to ensure a coherent AR system, we have to align the virtual and the real world which amounts to estimate the pose of the real camera. Thus, this issue has attracted a large scientific community. Therefore, many types of sensors have been considered: mechanical, ultrasound, magnetic, inertial, GPS, compass, gyroscope, and accelerometer. Nevertheless, the camera is the most used one.

A lot of researches have been conducted in this field. However, a few many reviews and surveys have been done in order to list and classify the proposed techniques. Teichrieb et al. [11] presented a review on online monocular marker-less augmented reality, dividing the approaches into two categories: model based (edge based, optical flow based and texture based) and structure from motion based (real time SFM, Mono SLAM). More recently, Marchand et al. [12] presented a survey on augmented reality describing the mathematical aspect of pose estimation techniques.

A survey of mobile AR is presented in [13] that describes the latest technologies and methods to improve runtime performance and energy efficiency for practical implementation. In the same context, we can find the history of mobile augmented reality in [14]. Rabbi and Ullah [15] presented a survey on AR challenges and tracking techniques.

The aim of this paper is to provide a technical classification of most of approaches for 3D pose estimation, we also focus on key-points-based techniques and present a reach comparison of both detectors and descriptors of the state of the art. The study given in this paper may help developers and researchers in the field of augmented reality.

The remain of this paper is organized as follow: section 2 describe the AR principle, section 3 presents the pose estimation techniques according to available data (3D or 2D), section 4 is dealing with features detection and description techniques, here we present a comparison considering computing time, recognition rate and memory space, finally we give a brief conclusion of this work.

## II. AUGMENTED REALITY PRINCIPLE

In order to achieve a coherent augmented scene, that combines both virtual and real worlds, we have to align the real and the virtual cameras. In other words, we have to assign to the virtual camera, the same properties (extrinsic and intrinsic) as those of the real camera. Thus, we need to determine in real time the position and the orientation of the camera for each frame in the real scene. The following figure (Fig. 1) illustrates the 2D-3D registration problem.

Let $R_w$, $R_C$, $R_{vw}$, $R_{vc}$ and $R_i$ respectively represent the real-world landmark, the camera landmark, the virtual world landmark, the virtual camera landmark and the image landmark. In order to get a coherent composition of the real and virtual world, the two real and virtual cameras should have the same position and the same parameters (focal, field of view (FOV), etc.) according to the reference points of the two real and virtual worlds ($R_w$, $R_{vw}$). Hence, the only unknown is the pose of the real camera relatively to the real-world landmark.

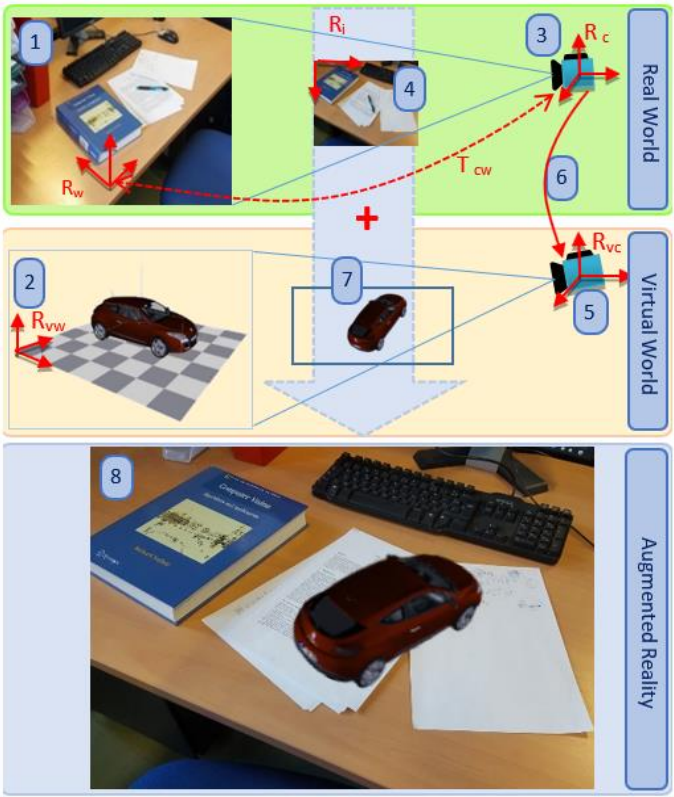

Fig. 1. General principle of registration process in augmented reality. 1) Real environment. 2) Virtual environment. 3) Camera. 4) Captured image. 5) Virtual camera. 6) Alignment of the virtual camera with the actual camera. 7) Projection space. 8) Augmented reality.

Let P be a point in the real world coordinate $(X_w, Y_w, Z_w)^T$ in $R_w$ and $(X_c, Y_c, Z_c)^T$ in $R_c$, the transformation from $R_w$ to $R_c$ is described as follows [16] (1):

$$\begin{pmatrix} X_c \\ Y_c \\ Z_c \end{pmatrix} = r \begin{pmatrix} X_w \\ Y_w \\ Z_w \end{pmatrix} + t = (r\ t) \begin{pmatrix} X_w \\ Y_w \\ Z_w \\ 1 \end{pmatrix} \ldots\ldots (1)$$

Where (r t) represents the transformation between the two landmarks (world and camera). This is defined by the translation vector (t) and the rotation matrix (r) of $R_w$ to $R_c$. Let Q be the perspective projection of P on the image plane. The coordinates of this projection can be calculated as follows [16] (2):

$$s \begin{pmatrix} u \\ v \\ 1 \end{pmatrix} = \underbrace{\begin{pmatrix} \alpha_u & 0 & u_0 \\ 0 & \alpha_v & v_0 \\ 0 & 0 & 1 \end{pmatrix}}_{A} \begin{pmatrix} X_c \\ Y_c \\ Z_c \end{pmatrix} = A \underbrace{(r\ t)}_{T} \begin{pmatrix} X_w \\ Y_w \\ Z_w \\ 1 \end{pmatrix} \ldots (2)$$

Where "A" is the matrix of intrinsic parameters ($\alpha_u$, $\alpha_v$: the ratio between the focal length and the horizontal and vertical size of the pixel, $u_0$, $v_0$: the intersection of the optical axis with the image plane) and T the matrix of extrinsic parameters. We assume that "A" is known, so that we obtain the following equation (3):

$$q = A^{-1}Q = TP \ldots\ldots\ldots (3)$$

As mentioned, in order to insert a virtual object in a real scene in a coherent way, we have to know the pose of the camera that we represent here by the matrix "T". Thus, if we have a set of points $P_i(X_i, Y_i, Z_i)$ and their projections $q_i(x_i, y_i)$, we can determine the transformation T.

We present in the following the different approaches that determine the pose of the camera, or in other words, to solve the following equation (4):

$$q_i = TP_i \ldots\ldots\ldots\ldots (4)$$

### III. POSE ESTIMATION IN AR

We illustrate in Figure 2 the different approaches allowing the pose estimation according to the available data (3D or 2D), from the P-nP problem to the SLAM [11], in addition to the planar scene. Fig. 2 presents a classification of pose estimation techniques.

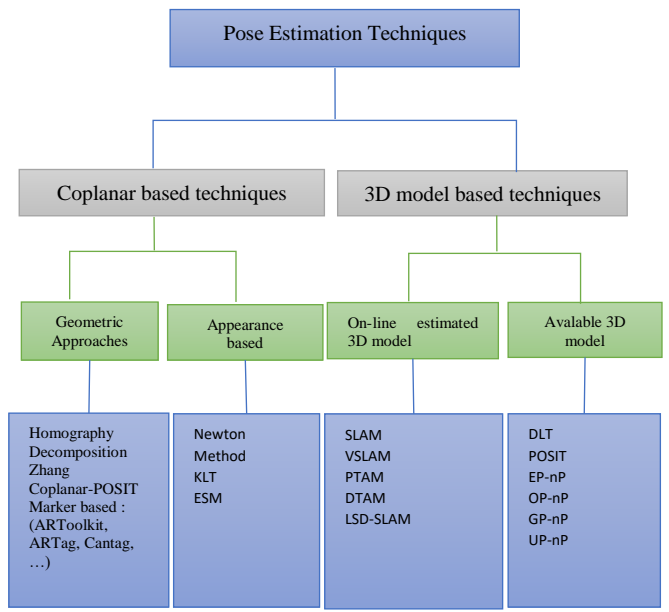

Fig. 2. Classification of the different approaches for 3D pose estimation in AR.

#### A. Pose Estimation based on 3D model

3D Pose can be estimated using a minimum of 3 points. Indeed, the pose can be represented by six parameters (3 angles of rotation and 3 translations), therefore 3 points would be sufficient to solve the equation (4), which corresponds to the problem P-3P (Perspective 3 Points).

Solving this problem comes down to two steps: The first step consists on estimating the $Z_{ic}$ for each point with respect to the $R_c$ reference via the cosine law theorem [17] using the triangle $CP_iP_j$ (with C is the origin of the reference $R_c$). Once we have the 3 points coordinates, the second step is to estimate the transformation T which makes it possible to carry out the passage from $R_m$ to $R_c$.

As a second alternative, the least squares method can be used. It gives an ambiguous solution and requires a fourth point to have a unique solution. This one is based mainly on the singular value decomposition SVD [12].

Kneip and al. [18] proposed a new solution to P-3P problem which calculates T directly in one step, without estimating the coordinates of the points with respect to the reference of the camera $R_c$. This is made possible by introducing the camera landmark $R_{c'}$ and the world landmark $R_{m'}$. Therefore, the projection of points from $R_{m'}$ to $R_{c'}$ reduces the problem to two conditions.

Although, P-3P approaches give solutions to pose estimation problem, but P-nP approaches give more accurate results by using more points. Quan and Lan [19] extended their P-3P algorithm to P-4P and then P-5P to finally reach P-nP. In the EP-nP approach [20], 3D point coordinates are expressed as a weighted sum of four virtual control points. The pose problem is then reduced to the estimation of the coordinates of these control points in the camera reference. This approach reduces computational complexity.

Direct Linear Transformation (DLT) is certainly the oldest one-step approaches [21]. Although not very accurate, this solution and its derivatives have largely been taken into account in AR applications.

P-nP is a non-linear problem. Among the solutions that take into account the non-linearity of the system, POSIT is an iterative approach proposed by Dementhon and Davis [22], the main idea consists on the use of an orthogonal projection system so that the problem becomes linear, then iteratively we return to the basic perspective projection system.

When N increases considerably, there is no solution with linear complexity for the problem P-nP. Possible solutions for this case include EPnP [20], OPnP [23], GPnP [24], and [25].

Other methods are based on tracking a model for pose estimation. The idea is to define a distance between the point of a contour in the image and the projection of the 3D line corresponding to the 3D model. The pose is estimated by minimizing the error between the selected points and the projected contours (Fig 3).

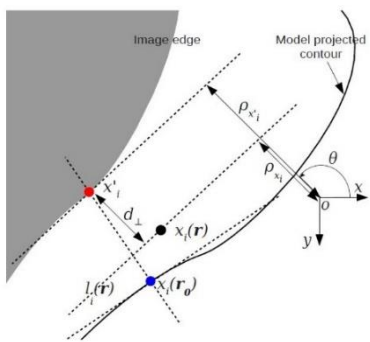

Fig. 3. Contour tracking for the pose estimation, extracted from [26]. From the initial pose $r_0$, a one-dimensional search is performed along the normal to the projected contour underlying the measurement point $x_i(r_0)$. And minimizing the distance d between the point $x'_i$ and the line $l_i(r)$.

Comport and associates [27] have proposed a tracking algorithm based on a 3D model. A nonlinear estimation of the pose is formulated using a virtual visual servoing approach.

The previous approaches are based on available 3D model. Other ones estimate at the same time the structure of the scene, and the pose of the camera. These approaches are called VSLAM (Vision based Simultaneous Localization and Mapping) [28] [29] [30].

Davison [28] used the extended Kalman filter for data integration. On the other hand, Eade and Drummond [29] used the particle filter. In such approaches, the data is sequentially integrated into the filter. The updates (camera position, speed, scene structure) are made sequentially. So, the number of estimated parameters increases with the map size.

Bundle Adjustment (BA) approaches estimate the movement of the camera by minimizing the error between the predicted points and the observed points, thus to build the map [31] [32].

Although some studies [33][31] have demonstrated the possibility of using SLAMs in AR, nevertheless this kind of approaches is lacking in terms of absolute localization and remains complex and expensive in computing time.

PTAM (Parallel Tracking and Mapping) [34] [35] dissociates tracking from mapping. Its main idea is to compute in parallel the map using BA technique and the pose using a tracking method. This approach is used in various application areas, particularly in AR [36] [37].

There are several works that are based on the use of a lot of cameras or other types of sensors, such as Kinect, like KinectFusion [38]. These kinds of systems make possible to directly determine the 3D position of the points. However, they require a slow learning and reconstruction step.

### B. Pose Estimation based on coplanar informations

Considering a planar scene simplifies the pose estimation problem. It's amount to camera motion estimation process. In this section we present the approaches based on extraction of 2D information from the image and the geometric information of a planar scene. The objective is to estimate camera move between two images instead of estimating the pose; the 3D model is therefore replaced by a reference image.

#### 1) Geometric Approaches

The objective is to estimate the camera 3D movement between two acquisitions using only the 2D information extracted from the two images. The homography between the two images is often used in this case [39], [40].

Let $x_1$ be a point of the image $I_1$ and $x_2$ a point of the image $I_2$ ($I_1$, $I_2$ two images of the same scene with a different view point). The two points ($x_1$, $x_2$) are related by the homography $H_1^2$ as follows (5) [12]:

$$x_2 = H_1^2 \, x_1 \quad \ldots\ldots\ldots\ldots\ldots\ldots(5)$$

$H_1^2$ can be estimated using the Direct Linear Transformation (DLT) algorithm [21]. The pose is then calculated by a homography decomposition [41].

The simplicity of this approach made its use in AR a standard. Thus, several coded target identification systems are based on this approach. The idea is to place different markers by their colors or shapes in the real environment. Based on the fact that the markers are known a priori we can estimate the 3D camera pose.

More recently, DeTone et al. [42] have considered the homography estimation problem between two images as a learning problem. They applied a convolutional neural network (CNN) to solve it. However, this type of neural network is not relevant in term of computation time.

*2) Appearance based Approaches*

Considering the 2D model as a reference image, the objective is to estimate the movement between the captured image and the reference image in pixel scale. Since the model is defined by a set of pixels, we must find their new positions in the image. Instead of using homography to determine pose, alignment can be defined directly as a minimization problem of dissimilarities or maximization of similarities between the appearance of the area of interest in the reference image and the area of interest in the captured image.

For example, if the appearance is defined as the pixels intensity belonging to a patch, the dissimilarity is considered as the SSD (Sum of Squared Differences) differences, which amounts to the KLT algorithm proposed in [46].

The tracking algorithm proposed by Benhimane & Malis [43] is based on the minimization of the SSD (Sum of Squared Differences) between a given model and the current image by applying the ESM algorithm (Efficient Second Order Minimization) which has the same convergence properties as Newton's method, but with a faster computation time.

IV. FEATURES DETECTION AND DESCRIPTION

Although research related to markers is still active, recent advances in computer vision and the development of key-points matching methodologies make AR reaching a new level maturity. We present in the following the different techniques of detection and description of image features.

*A. Features Detection*

Detecting stable points of interest is an essential step for any point-based computer vision process. Recently, corner detection becomes the most widely used technique because of its good performance in terms of repeatability and processing time.

A corner is defined as the intersection point of two lines or outlines. Mathematically, it refers to the point where there are two dominant but different gradient orientations. Different approaches of corner detection can be classified into three categories [44]:

1. **Gradient-based corner detection:** These techniques are based on gradient calculation, such as, Harris [45], KLT [46], and Shi-Tomasi [47]. These techniques are robust but expensive in computing time.

2. **Contour-based corner detection:** These techniques study the shape of contours to identify corners. Namely DoG-curve detectors [48], ANDD [49] and Hyperbola fitting [50]. However, these techniques are very sensitive to noise.

3. **Model-based corner detection:** These techniques are based on the comparison of pixels around a model. SUSAN [51] and FAST [52] use this approach. Recently, models have been combined with machine learning techniques, such as decision trees, for faster detection of corners. Such as the AGAST detector. These techniques are the most used for real-time applications.

As the first model-based corner detector, Fast Accelerated Segment Test (FAST) represents a breakthrough in the evolution of corner detectors. It is based on the Accelerated Segment Test (AST) principle which is a modification of the SUSAN detector.

The FAST-ER detector [53] is an improved version of FAST, increases the thickness of the circular model to increase the stability of the detected corners; however, it has become slower compared to FAST.

The AGAST detector [54] is also based on the AST technique. It has considerably improved the construction and use of the decision trees used in the AST, which makes it more efficient than FAST in terms of time calculation and repeatability.

Recently, Bellarbi et al. proposed MOBIL_Detector [55] which hybrids AGAST detector and the Shi-Tomasi measurement in order to obtain more stable key-points.

We compared the computing time of 5 popular detectors. Table 1 illustrates this comparison.

TABLE 1. Average time to detect 500 points of interest for MOBIL_Detector, SURF, ORB, BRISK and AGAST detectors.

| Detector | Time (ms) |
|---|---|
| SURF [56] | 72.63 |
| ORB Detector [57] | 8.37 |
| BRISK Detector [58] | 4.12 |
| AGAST [54] | 1.87 |
| MOBIL_Detector [55] | 3.76 |

As shown in Table 1, MOBIL_Detector, AGAST and BRISK Detector are faster than ORB detector, and much better than SURF. Which makes them more suitable for real-time applications like AR.

*B. Features Description*

A descriptor is a function applied to the patch in order to describe it, in an invariant way for any changes to the image (eg, rotation, lighting, noise, etc.).

The common pipeline for using descriptors is:

1. Select regions (patches) around the detected key-points in the image. These patches are square or circular shapes depending on the properties of the descriptor to be applied.
2. Describe each region (the patch) as a feature vector, using this descriptor.
3. Calculate the distance between vectors using a similarity measure.

In the state of the art, most of works focuses on the description of key-points. Those description techniques (descriptors) have been grouped into two main families, Floating Point Descriptors, and Binary Descriptors. We present in the following two comparative tables of the most known descriptors.

The first table (Table 2) illustrates the computing time of some known descriptors, we calculated the average time of a description of a patch for each of these descriptors. Thus, we noticed that the binary ones are more suitable for real-time applications (at least 15 frames per second), than the floating-point descriptors.

We note that the description is made on 500 points / frame and the number of frames per second is calculated according to the time of the description only (without adding the time of detection and matching).

TABLE 2. Description mean time of a patch (ms) and the number of frames / sec.

| Descriptors | Description mean time of a patch (ms) | Number of frames per second (description of 500 points per frame) |
|---|---|---|
| SIFT [59] | 3.121 | 0.64 |
| SURF [56] | 1.488 | 1.34 |
| LDA-HASH [60] | 4.21 | 0.47 |
| BRISK [58] | 0.072 | 27.77 |
| FREAK [61] | 0.094 | 21.27 |
| ORB [57] | 0.146 | 13.69 |
| LDB [62] | 0.139 | 14.38 |
| LATCH [63] | 0.437 | 4.57 |
| MOBIL [64] | 0.127 | 15.74 |
| MOBIL_2B [65] | 0.136 | 14.70 |
| POLAR_MOBIL [55] | 0.107 | 18.68 |

We present in Table 3, a comparison of more than 30 descriptors from the state of the art with their detectors. We have classified these descriptors according to certain criteria that we judged necessary for the implementation of an augmented reality application.

We evaluated the descriptors according to each criterion, namely *computation time*, *recognition rate* and *memory space*, using a scale from 1 to 5 (from + to +++++) shown as follows:

| Computing time: | Recognition rate: | Memory: |
|---|---|---|
| +: Very slow. | +: Less robust. | +: Voluminous. |
| +++++: Very speed. | +++++: Robust. | +++++: Lightweight. |

According to Table 3, we noticed that new descriptors based on deep learning such as DeepBit [66], DELF [67] and LIFT [68] give better results in terms of recognition rate. However, their major disadvantage is the calculation time, similarly to traditional floating-point descriptors such as SIFT [59], GLOH [69], LDE [70] or DAISY [71].

On the other hand, we noticed that the binary descriptors are better than the other two families in terms of memory and computing time. However, their robustness and their discriminative and distinctive powers are considerably limited. Except, some robust binary descriptors such BRISK, FREAK, POLAR_MOBIL and BOLD, which are suitable for using in real-time augmented reality applications.

V. CONCLUSION

This paper presents a global vision of the pose estimation problem for augmented reality applications. We first based on the geometrical aspect of the pose estimation by presenting the different methods that make it possible to geometrically answer this problem, then we approached the approaches that integrate motion estimation. We classified these approaches according to the available information: 3D model or planar scene.

Then, we presented the extraction and description of features and we presented a comparison of different descriptors. According to the conducted comparison, we found that the recent binary descriptors are the most suitable for such augmented reality applications, thanks to their low computing time and memory consumption.

TABLE 3. Descriptors Comparison

| Descriptor | Suggested Detector | Type | Computing time | Recognition rate | Memory space |
|---|---|---|---|---|---|
| SIFT [59] | DoG | Float | + | +++++ | ++ |
| PCA-SIFT [72] | DoG (SIFT) | Float | ++ | ++++ | +++ |
| GLOH [69] | DoG (SIFT) | Float | + | +++++ | ++ |
| SURF [56] | Determinant of Hessian | Float | + | ++++ | ++ |
| LDE [70] | DoG | Float | ++ | +++++ | +++ |
| CONGAS [73] | LoG | Float | ++ | ++++ | ++ |
| Daisy [71] | DoG (SIFT) | Float | + | +++++ | ++ |
| BRIEF [74] | SURF Detector | Binary | +++ | +++ | ++++ |
| ORB [57] | FAST | Binary | ++++ | ++++ | ++++ |
| BRISK [58] | AGAST+FAST | Binary | +++++ | ++++ | ++++ |
| FREAK [61] | BRISK Detector | Binary | +++++ | ++++ | ++++ |
| ALOHA [75] | SURF Detector | Binary | +++ | ++++ | +++ |
| LDA-HASH [60] | DoG (SIFT) | Binary | + | +++++ | + |
| KAZE [76] | Hessian Matrix + Scharr filter | Float | ++ | +++ | +++ |
| D-BRIEF [77] | DoG | Binary | ++ | ++++ | ++ |
| BinBoost [78] | DoG | Binary | ++ | ++++ | ++ |
| A-KAZE [79] | KAZE Detector | Binary | ++++ | ++++ | +++ |
| BRIGHT [80] | DoG | Binary | ++ | +++ | ++ |
| LDB [62] | DoG (SIFT) | Binary | +++ | ++++ | +++ |
| OSRI [81] | DoG / Hessian /Harris-Affine | Binary | +++ | ++++ | ++++ |
| BAMBOO [82] | Not specified | Binary | +++ | +++ | +++ |
| USB [83] | DoG | Binary | +++ | ++++ | ++++ |
| PRO [84] | Multi-scale Harris | Binary | +++ | ++++ | ++++ |
| MOBIL [64] | FAST | Binary | ++++ | ++++ | ++++ |
| BSIFT [85] | SIFT | Binary | +++ | ++++ | +++ |
| BOLD [86] | Haris-Laplace | Binary | ++++ | ++++ | ++++ |
| MOBIL_2B [65] | FAST | Binary | ++++ | ++++ | ++++ |
| Deep Hashing [87] | Full Image | Binary | ++ | +++++ | +++ |
| LATCH [63] | Multi-scale Harris | Binary | ++ | ++++ | +++ |
| DeepBit [66] | Full Image | Binary | ++ | +++++ | +++ |
| DELF [67] | Convolutional neural network | Float | + | +++++ | ++ |
| LIFT [68] | Convolutional neural network | Float | ++ | +++++ | + |
| POLAR_MOBIL [55] | MOBIL_DETECTOR | Binary | ++++ | ++++ | ++++ |